\documentclass[letterpaper, 10 pt, journal, twoside]{ieeetran}

\IEEEoverridecommandlockouts                              

\usepackage{outlines}
\usepackage{amsmath} 

\usepackage[noabbrev]{cleveref}
\usepackage{booktabs}
\usepackage{multirow}
\usepackage{graphicx}
\usepackage{subcaption}
\usepackage[export]{adjustbox}

\usepackage{soul}

\usepackage{dblfloatfix}

\usepackage{comment}
\usepackage{url}

\title{\LARGE \bf
Panoptic One-Click Segmentation: Applied to Agricultural Data
}

\author{Patrick Zimmer, Michael Halstead, Chris McCool
\thanks{Preprint, accepted on February 18, 2023 for publication in IEEE RA-L. DOI: 10.1109/LRA.2023.3254451}
\thanks{© 2023 IEEE.  Personal use of this material is permitted.  Permission from IEEE must be obtained for all other uses, in any current or future media, including reprinting/republishing this material for advertising or promotional purposes, creating new collective works, for resale or redistribution to servers or lists, or reuse of any copyrighted component of this work in other works.}
\thanks{This work was partly funded by the Deutsche Forschungsgemeinschaft (DFG, German Research Foundation) under Germany’s Excellence Strategy - EXC 2070 – 390732324.
The authors would like to thank the German Federal Ministry of Food and Agriculture (BMEL) for funding the WeedAI project. In particular, we would like to thank the Rhenish beet growers association Rheinischer-R\"{u}benbauer-Verband e.V. (Bonn, Germany) for support in the project.}
\thanks{The authors are with the University of Bonn, Germany; Agricultural Robotics Department
{\tt\small \{patrick.zimmer, michael.halstead, cmccool\}@uni-bonn.de}}
\thanks{Our code is released at github.com/Agricultural-Robotics-Bonn/UniBonn-Agrobot-PanopticOneClick}
}

\begin{document}

\maketitle
\IEEEpeerreviewmaketitle

\begin{abstract}
In weed control, precision agriculture can help to greatly reduce the use of herbicides, resulting in both economical and ecological benefits.
A key element here is the ability to locate and segment all the plants (crop \& weed) from image data.
Modern instance segmentation techniques can achieve this, however, training such systems requires large amounts of hand-labelled data which is expensive and laborious to obtain.
Weakly supervised training can help to greatly reduce labelling efforts and costs.

In this paper we propose panoptic one-click segmentation, an efficient and accurate offline tool to produce pseudo-labels from click inputs and thereby reduce labelling effort when creating novel datasets.
Our approach jointly estimates the pixel-wise location of all $N$ objects in the scene, compared to traditional approaches which iterate independently through all $N$ objects.
This results in a highly efficient technique with greatly reduced training times.
Using just 10\% of the data to train our panoptic one-click segmentation approach yields 68.1\% and 68.8\% mean object intersection over union (IoU) on challenging sugar beet and corn image data respectively, providing comparable performance to traditional one-click approaches while being approximately 12 times (an order of magnitude) faster to train. 
We demonstrate the practical applicability of our system by generating pseudo-labels from click annotations for the remaining 90\% of the data.
These pseudo-labels are then used to train Mask R-CNN, in a semi-supervised manner, improving the absolute performance (of mean foreground IoU) by 9.4 and 7.9 points for sugar beet and corn data respectively, demonstrating the potential of our approach to rapidly annotate challenging data.
Finally, we show that our panoptic one-click segmentation technique is able to recover missed clicks during annotation outlining a further benefit over traditional approaches.

\end{abstract}

\section{Introduction}

Weed control in agricultural settings is an important part of ensuring crop health and subsequently high yield.
For automated systems this is heavily influenced by technological developments from fields such as robotics and computer vision.
Vision based systems in combination with modern deep learning (DL) algorithms have become increasingly popular in this domain~\cite{Hasan2021, Zhang2022}.
Unfortunately, while these techniques can achieve high accuracy in the agricultural domain, they generally require large amounts of labelled data for training.
One approach to alleviate the associated costs with acquiring this labelled information is weak labelling.

\begin{figure}[t!]
\centering
    \includegraphics[width=.40\textwidth]{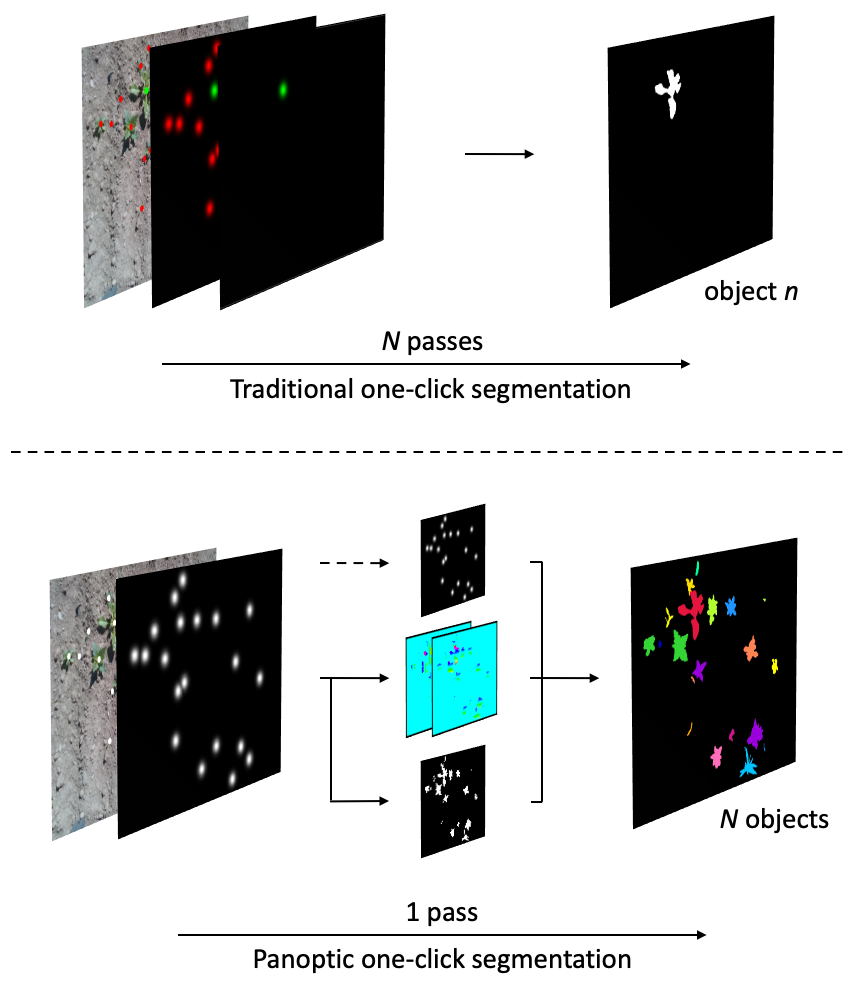}
    \caption{Comparison of a traditional one-click segmentation approach (top) and our novel panoptic one-click segmentation approach (bottom).
    The traditional approach takes a \textit{positive} click and optionally \textit{negative} clicks as input and segments each of the $N$ objects in $N$ separate forward passes.
    By comparison, our approach uses a data representation from \cite{Cheng2020} and jointly resolves all $N$ clicks and objects in 1 forward pass.
    Clicks in RGB images and transform maps are highlighted and colored only for visualization.}
    \vspace{-16pt}
\label{fig:hero}
\end{figure}

Weak labels are considered any label that is less informative than fully annotated labels (i.e. pixel-wise labels for segmentation).
For weak segmentation approaches~\cite{Xu2016,Xu2017,Maninis2018,Zhang2020c,Forte2020}, the primary goal is to obtain fully annotated outputs (or as close to full as possible) from weak priors such as bounding boxes or even single instance clicks.
These methods can be used as a tool to assist interactive segmentation and speed up annotation or rendering tasks~\cite{Xu2016,Forte2020}, or as a base to provide labels for fully supervised systems~\cite{Dai2015,Khoreva2017}.
For these fully supervised systems the generated pseudo-labels can replace expensive manual pixel-wise labels for training segmentation networks, typically with the goal to achieve comparable performance to manual pixel-wise approaches.
Typically, semi-supervised techniques combine both manually annotated ground truth with pseudo-labels from weak learning approaches.
This creates large datasets with minimal expense associated with the labelling (i.e. single click per object compared to pixel-wise segmentation) for training DL algorithms.

In this paper we present an efficient and accurate method for weak labelling enabled by panoptic segmentation.
We introduce panoptic one-click segmentation, a novel offline tool to produce pseudo-labels from a single click (annotation) per object in the image.
This approach greatly simplifies the creation of novel datasets (e.g. for semi-supervised learning).
We compare our panoptic one-click segmentation approach to a traditional click segmentation method~\cite{Xu2016} adapted to this setting.
This traditional approach learns to produce a segmentation mask for an object based on the following input: an RGB image, one click for the object of interest and potentially one click for all of the other objects in the image.
This is a computationally inefficient technique as it processes each of the $N$ objects individually, this leads to long training times.
We overcome this inefficiency by employing panoptic segmentation~\cite{Cheng2020} to jointly resolve the segmentation mask for all $N$ objects simultaneously, as depicted in~\Cref{fig:hero}.

We evaluate our panoptic-based approach on two challenging arable farming (crop/weed) datasets from sugar beet and corn fields (see \Cref{sec:experiments:datasets}). 
Furthermore, we demonstrate the practical applicability of this one-click segmentation approach to train Mask R-CNN using just 10\% of the manually annotated ground truth; the remaining 90\% of the data is generated as pseudo-labels using our approach.
In doing so, we make the following contributions:
\begin{itemize}
    \item we propose panoptic one-click segmentation, a novel offline tool for producing pseudo-labels using just one click per object which yields competitive performance to a traditional approach while being an order of magnitude faster to train; and
    \item we demonstrate the practical applicability of our approach to train Mask R-CNN using only 10\% of the manually annotated ground truth with the remaining 90\% of the data consisting of pseudo-labels generated using our panoptic one-click segmentation approach; and
    \item we demonstrate the potential for our panoptic one-click segmentation system to accurately recover missed clicks from the annotation process.
\end{itemize}

\section{Related work}

Accurate segmentation of plants in the field enables a range of tasks, from field or crop monitoring~\cite{Halstead2021} through to precise weeding~\cite{Ahmadi2022a}.
To achieve this, current state-of-the-art approaches have to be trained in a supervised manner which requires a large number of images to be annotated with a label for each pixel, however, these pixel-wise annotations are expensive and labour intensive to acquire \cite{Zhang2022}.
An alternative is to perform semi-supervised training by using sparse labels for each object~\cite{Dai2015,Khoreva2017}, we refer to this as weak labelling.

In this section we review literature with a primary focus on weakly labelled segmentation, and then briefly outline semi-supervised learning, as well as its use within the agricultural domain.
Finally, as our novel approach leverages panoptic vision for one-click segmentation we briefly review panoptic segmentation.

\subsection{Weakly Labelled Segmentation} 
\label{sec:related:weak_seg_baselines}

Weakly labelled segmentation is a method where segmentation networks are trained to obtain pseudo-labels from less than pixel-wise annotations.
A major benefit of weak labelling is that sparse labels can be obtained much quicker than annotating each pixel of that object.
There are a range of approaches for weak label based segmentation which can differ in terms of the target setting~\cite{Xu2016,Majumder2021}, type and amount of user input~\cite{Maninis2018,Majumder2021}, input processing~\cite{Xu2016,Lin2021a}, or network structures \cite{Xu2016,Forte2020}.
Despite this diversity, many of these weak labelling approaches share similarities while using different input types as is the case of~\cite{Xu2016} which uses clicks as input whereas~\cite{Xu2017} use bounding boxes as their input.
And yet, even using clicks is a diverse area as the clicks can be from anywhere in the object~\cite{Xu2016} or there can be a series of clicks placed on specific locations to contain the object~\cite{Maninis2018}.
Therefore, weak labelling is becoming an extensive area of research with many varying methods for obtaining pseudo-labels.

One of the earliest deep learning-based approaches for click-based segmentation was proposed in 2016 by Xu et al.~\cite{Xu2016}.
Each click was transformed to a map using the Euclidean distance of a pixel from the click, we term this a click transform map.
The click transform map was then used as a fourth channel in addition to the image (three channels for R, G, \& B) as input for a fully convolutional segmentation network (FCN)~\cite{Long2015}.
Such an approach can be used for a single click or for multiple clicks when used in the context of interactive segmentation; the clicks could be both positive (in the object) and negative (outside the object) clicks.

Subsequent works typically build upon this initial work and its click map guidance approach.
Maninis et al.~\cite{Maninis2018} made use of clicks on the extremity of objects, termed DEXTR.
Aside from using clicks in predefined positions, this work also made use of an improved network architecture consisting of an encoder-decoder structure rather than an FCN structure.
Further, their method uses a 2D Gaussian transform instead of the Euclidean distance.
A similar approach was proposed in~\cite{Zhang2020c}, likewise with predefined click positioning consisting of the object center as well as the opposite corners of an enclosing bounding box (which are negative clicks).

Other recent work has considered clicks explicitly as a sequence in an interactive segmentation paradigm similar to \cite{Xu2016}.
Mahadevan et al.~\cite{Mahadevan2018} incorporated masks from previous iterations for refinement as an additional input channel, using DeepLabV3+~\cite{Chen2018} and obtaining click maps by using a 2D Gaussian transform.
Forte et al.~\cite{Forte2020} also exploited the click sequence by treating the sequential clicks in a separate processing stream.
They used a UNet~\cite{Ronneberger2015} style architecture with further adaptions and aimed at obtaining the highest quality segmentation (finest possible detail).
In all of these approaches the authors generate pseudo-labels which could be used for training semi-supervised systems.

\subsection{Semi-supervised Applications}

Semi-supervised segmentation is any approach that trains using a proportion of pseudo-labels (not manually annotated) or on a percentage of data that is less than the fully annotated data.
In this case we outline techniques that are based on pseudo-labels produced by weak labelling techniques.
Using their weak segmentation method on objects in the PASCAL VOC 2012 dataset~\cite{pascal-voc-2012}, Dai et al. \cite{Dai2015} generate pseudo-labels from bounding box priors.
They find that performance is not impacted when training semantic segmentation with 90\% of pseudo-labels and only 10\% of hand labelled data.
Similarly, \cite{Khoreva2017} was able to achieve 95\% and 99\% of the fully supervised results on PASCAL and PASCAL/COCO~\cite{Lin2014} (combined) datasets respectively.
In our work, we use a similar data split, generating pseudo-labels from models trained on 10\% of our datasets, predicted on the remaining 90\%.

\subsection{Agricultural Applications}

There have been few attempts at applying weak or semi-supervised learning techniques to the agricultural domain.
In viticulture~\cite{CasadoGarcia2022} Casado-Garcia et al. compare various approaches to leverage label information from just 20\% of their data and generate pseudo labels for the remaining 80\% unlabelled data which they combine to train semi-supervised semantic segmentation.
In~\cite{Ciarfuglia2022} Ciarfuglia et al. use bounding boxes obtained through a grape detector trained on a source dataset to then produce pseudo-labels for other data which are then applied to train an instance segmentation network.
To our knowledge, there is no prior work that uses click annotations to obtain pseudo-labels as well as use those for semi-supervised instance segmentation within the agricultural domain.

\subsection{Panoptic Segmentation}

Panoptic segmentation is a technique that aims to combine both semantic and instance segmentation.
It assigns each pixel in an image to either a \textit{stuff} or \textit{things} category~\cite{Kirillov2019}.
The \textit{stuff} category is for regions of similar texture or material such as grass, sky, road, or the ``background'' whereas the \textit{things} category relates to countable objects such as plants and fruit.
Additionally, it discriminates between instances of \textit{things} to enable the association of each pixel to a particular instance.
In this work we use Panoptic-Deeplab~\cite{Cheng2020} which uses a three head technique: semantic segmentation; center regression; and center offset.
At inference time the results of these three heads are used to produce the panoptic segmentation output using post-processing techniques.

\section{Proposed Approach}

We propose panoptic one-click segmentation which is a novel offline system to produce pseudo-labels from minimal user input.
In particular, we use a single click per object to produce pseudo-labels which can then be used to train other systems (e.g. in a semi-supervised learning pipeline).
Compared to traditional one-click segmentation which performs a forward pass for each of the $N$ objects in the image, panoptic one-click segmentation jointly estimates the pixel-wise location of all $N$ objects in a single forward pass.
Jointly processing all $N$ objects in a single pass reduces training and inference time by an order of magnitude.

We initially describe the traditional one-click segmentation approach, including a variant in which we incorporate negative instance clicks.
We then describe our panoptic one-click segmentation approach which jointly estimates the pixel-wise location of all objects in the scene.
Finally, we outline how this panoptic segmentation is able to recover missed or lost clicks during the annotation phase.

\subsection{One-click segmentation baseline} \label{sec:proposed:baseline}

\begin{figure}[t!]
 \vspace{4pt}
 \centering
    \includegraphics[width=.46\textwidth]{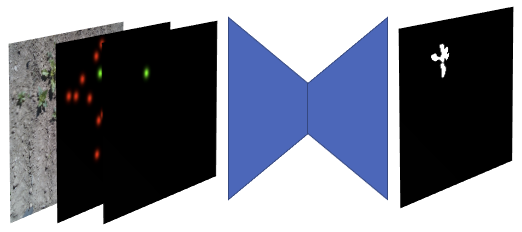}
    \caption{Overview of the baseline one-click semantic segmentation that we use.
    It consists of an encoder-decoder structure (in blue) which outputs all the pixels associated to one object.
    This is based on an input consisting of the RGB image and a click transform map for the corresponding positive click (in green) as a fourth channel.
    Optionally, we extend this system by using all $N$ object clicks, by adding the $N-1$ negative clicks (in red) on other objects within a second click map. 
    Click maps are colored only for visualization purposes.}
     \label{fig:standard}
     \vspace{-16pt}
\end{figure}

The baseline for our one-click object segmentation is based on the concept of~\cite{Xu2016}.
A click, $C_{(h,w)}$, is represented within a map which has the same size as the original image space $H\times{W}$.
Following more recent approaches \cite{Maninis2018,Lin2020}, we replace the originally used Euclidean distance transform and use a 2-D Gaussian instead, here with a standard deviation of 8.
This map is given as an extra input to a standard semantic segmentation network resulting in a four channel input RGB plus click transform  map.

When there are multiple objects (clicks) in the image the above procedure has to be applied iteratively.
At both training and inference time, each of the $N$ objects is treated separately such that $N$ semantic segmentation maps are produced. 
We refer to this approach as the standard one-click baseline.
A disadvantage of this approach is that it greatly increases the computational requirements as the same image is processed $N$ times.

\begin{figure*}[t!]
    \vspace{8pt}
    \centering
    \includegraphics[width=.8\textwidth]{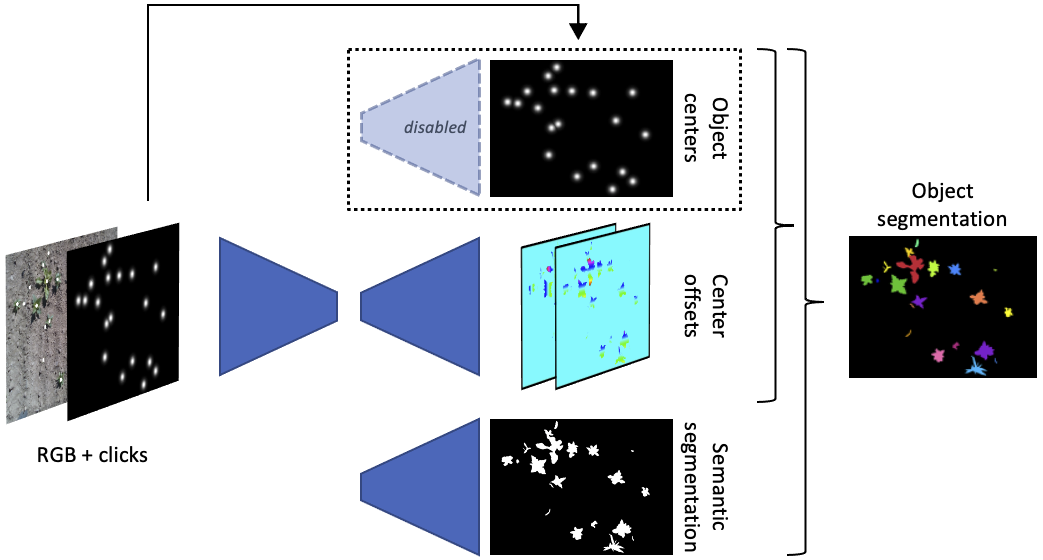}
    \caption{Overview of our panoptic one-click segmentation system which is based on panoptic deeplab~\cite{Cheng2020}.
    It takes as input the RGB image (3 channels) and a click transform map (clicks) as the fourth channel.
    Our system consists of two heads, one to estimate the vertical and horizontal offset to the center (offsets) and one for semantic segmentation.
    We disable the center estimation head and use the click centers from the user instead.
    The final output is instance-based object segmentation and is resolved as per~\cite{Cheng2020}.}
    \label{fig:scheme}
    \vspace{-16pt}
\end{figure*}

While this iterative approach is computationally expensive it does include the potential to be enhanced.
As all $N$ objects are annotated using a single click, for each positive click the remaining $N-1$ clicks from the other objects can be considered negative clicks.
These negative clicks are able to provide further information about the scene that a single positive click technique is not able to provide.
We encode all positive and negative clicks into a combined, second click map using the same Gaussian encoding as for the standard one-click approach. 
This full click transform map is then given to the network as the fifth channel in the input.
An overview of this system is provided in \Cref{fig:standard}.

\subsection{Panoptic one-click segmentation} \label{sec:proposed:panoptic}

We propose a panoptic one-click segmentation approach which jointly resolves the location of all objects in a single pass.
Our novel approach is based on Panoptic-Deeplab~\cite{Cheng2020} which we adapt to perform one-click-per-object segmentation for all $N$ objects within an image.
Panoptic-Deeplab produces instance segmentation labels using three output representations, depicted in~\Cref{fig:scheme}: (1) an object center map, (2) a center offset map, and (3) a semantic segmentation map.
The existence of an object (plant) is defined by a valid center location in the object center map.
The offset of every object (plant) pixel to its center location is estimated in the offset map and used to assign each pixel to the closest center location; the details of this post-processing algorithm are described in~\cite{Cheng2020}.
The segmentation map is a pixel-wise map of \textit{things} which in this case are the plants and \textit{stuff} which is everything else (e.g. background). 
Combining all three maps provides the instance-based segmentation mask.

For our panoptic one-click segmentation system only two heads are predicted by the network, these being the semantic segmentation map and the center offsets.
This is because the object centers are supplied by the annotator as a click map and so do not need to be predicted by the network.
Instead, the click map is provided as an extra input channel and is also provided as the object center locations as depicted in~\Cref{fig:scheme}.

\subsection{Panoptic system to recover from missing clicks} \label{sec:proposed:panoptic_miss} 
A potential advantage of panoptic one-click segmentation is the ability to estimate the location of object centers (e.g. clicks).
This would allow us to deal with potential annotation errors such as missing clicks.
To explore this possibility, we re-introduce the object center location estimation as the third head in the network. 
The input clicks are still used as the fourth channel at the input of the network, however, the object locations are now fully estimated by the network.
This means that we no longer use the clicks from the annotators in the post-processing stage.
A potential downside of this approach is that the center locations given by the user are only used as the input to the network.

\section{Experimental Setup}
We perform all experiments on two challenging agricultural weeding datasets.
Images typically contain multiple, frequently overlapping plant instances from various species and sizes.
Here, we describe both datasets in more detail and further provide information about input clicks, network implementations and evaluation metrics.
We implement our approach using PyTorch and all experiments were conducted on a single A6000 graphical processing unit (GPU).

\subsection{Datasets} \label{sec:experiments:datasets}

\begin{figure}[t!]
\vspace{8pt}
\centering
    \begin{tabular}{cc}
        \includegraphics[width=.23\textwidth]{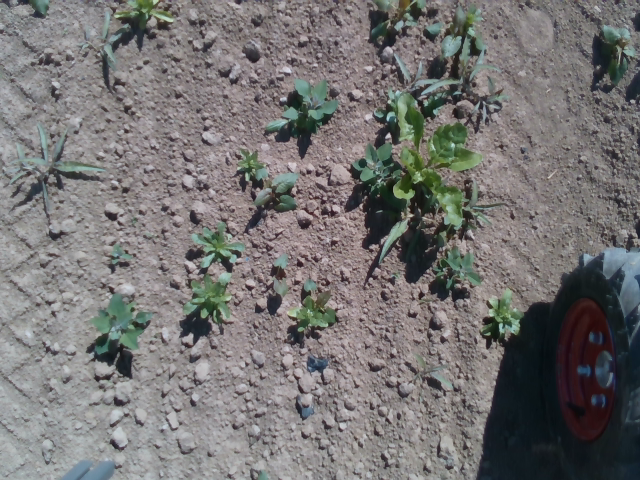} & \includegraphics[width=.23\textwidth]{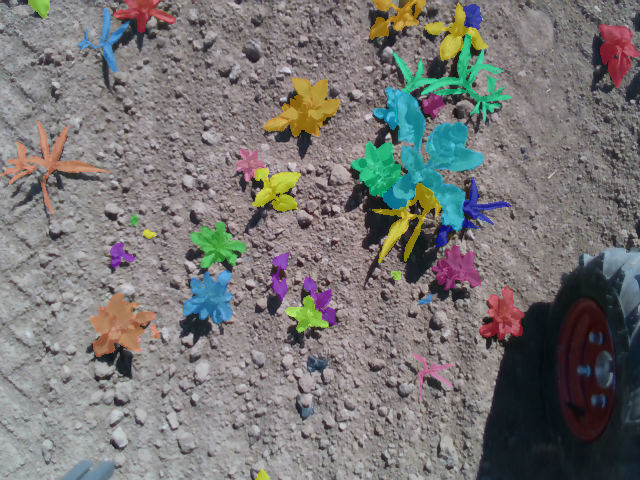} \\
        \includegraphics[width=.23\textwidth]{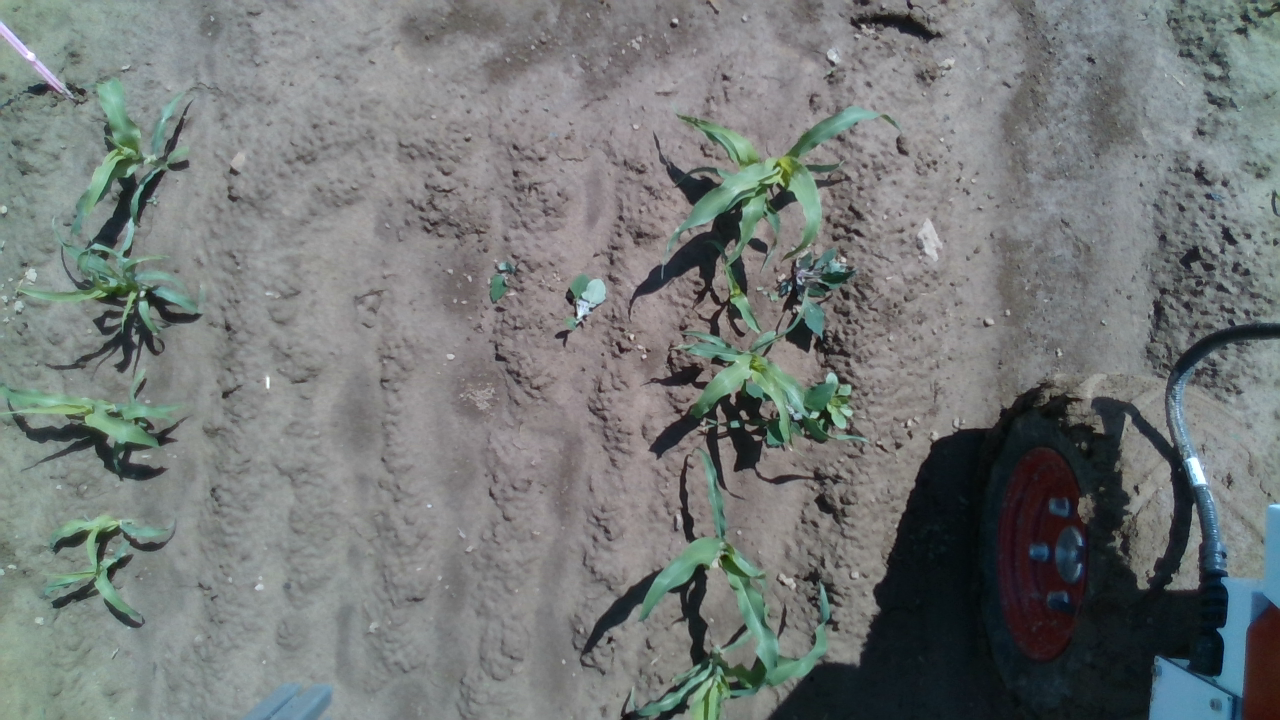} & \includegraphics[width=.23\textwidth]{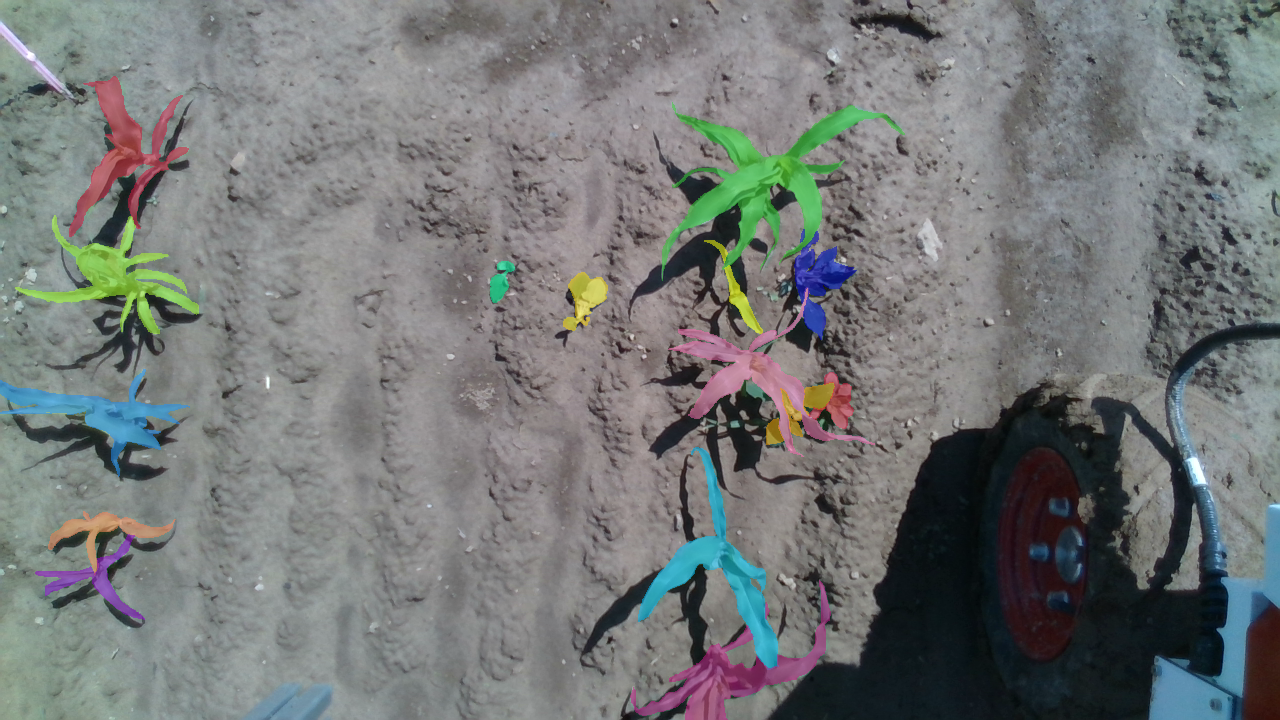}
    \end{tabular}
    \caption{Example data for both datasets \textit{SB20} (top) and \textit{CN20} (bottom). Left are plain RGB images and right additionally shows instance annotations.}
    \label{fig:dataset_examples}
    \vspace{-16pt}
\end{figure}

For all experiments we use two datasets that we refer to as \textit{SB20}~\cite{Ahmadi2021,Smitt2022,Halstead2021} (sugar beet 2020) and \textit{CN20}~\cite{Ahmadi2022a} (corn 2020).
Both datasets contain instance-level manually annotated data for both training and inference and contain varying weed species, sizes, and densities (see \Cref{fig:dataset_examples}).

These two datasets resemble typical arable farming settings.
The two crops, sugar beet and corn, are common crops (in Europe) and are highly variable in appearance.
Sugar beet has thick leaves and rosette-like shapes while corn consists of thin, long and irregularly twisted leaves.
The weeds present in both datasets (6 species for \textit{SB20} and 7 species for \textit{CN20}) likewise show strong diversity in appearance.
Both datasets are captured on a robot designed for crop monitoring and weed management~\cite{Ahmadi2022a} using a nadir Intel RealSense D435i camera.
\textit{SB20} consists of 71 training ($\sim$50\%), 37 validation ($\sim$26\%) and 35 evaluation ($\sim$25\%) images with a resolution of $640\times480$ and a total of 1424 plant instances in the training set.
\textit{CN20} consists of 150 training ($\sim$64\%), 40 validation (17\%) and 45 evaluation ($\sim$19\%) images with a resolution of $1280\times{720}$ and a total of 1781 plant objects for training.
\textit{CN20} is resized to $704\times416$ similarly to~\cite{Ahmadi2021}.

For our one-click segmentation experiments, we manually sub-sampled images from the training sets to contain roughly 10\% of the original object annotations (157 and 154 for \textit{SB20} and \textit{CN20}, respectively).
For our semi-supervised experiments, we use our one-click segmentation models to create pseudo-labels to replace the original annotations in the remaining 90\% of training data.

\subsection{Click inputs}

In both \textit{SB20} and \textit{CN20} dataset annotations, where visually possible, we also include keypoint/stem locations of the plants, which are used as our click locations.
However, due to the nuance of the datasets a small number of plants (63 and 30 instances for \textit{SB20} and \textit{CN20}, respectively) do not contain this keypoint location.
This is due to a number of complicating factors, such as, being on the border of the image or occlusion by other instances.

If the keypoint is not included we perform a two step pipeline to allocate a center point within the plant's binary mask.
First, we try using the center of mass of the binary mask to locate the click, if this is located outside of the segmentation map we disregard it and move to the second step.
We instead continuously perform binary erosion until complete disappearance of the object and randomly select one of the points remaining within the penultimate iteration.
This process ensures that the click is located within the object region.
Finally, to replicate the uncertainty that a human operator might annotate with, we add random noise of $\pm10$ pixels to the click positions when training models (guaranteeing they remain inside the object).

\subsection{Implementation and metrics}

\subsubsection{Standard one-click segmentation}
For our standard one-click models we train a UNet architecture \cite{Ronneberger2015}, using a batch size of 3 and a fixed learning rate of 0.0001.
We train from scratch for 1500 epochs, which we empirically found was the convergence point.
To tackle class imbalances in our datasets, we use a class weighted cross entropy loss similarly to~\cite{Milioto2020,Smitt2022}, that gives a higher loss to classes with fewer samples.
We determine class weights $w_c$ as,
\begin{equation}
\label{eq:loss}
    w_c = 1/\log \left( \frac{a_c}{a_{bg}}+1.02 \right),
\end{equation}
where $a_c$ is the area in pixels of class $c$ and $a_{bg}$ is the background area.
For evaluation, we report mean object IoU (mIoU) performance over the evaluation sets.

\subsubsection{Panoptic one-click segmentation}
\label{sec:experiments:panoptic}
For our panoptic system, we replace the original DeepLab architecture from \cite{Cheng2020} and replace it with UNet in order to stay comparable to our previous experiments.
Here, we use a batch size of 1 due to our small dataset sizes, and a fixed learning rate of 0.001.
The network is trained for 500 epochs as we found this to be the convergence point.
For training the semantic segmentation head, we use the same class weighted cross entropy loss as previously described.
Again, we report mean object IoU (mIoU) performance over the evaluation sets.

\subsubsection{Semi-supervised instance segmentation}
In our semi-supervised experiments, we train Mask R-CNN \cite{he2017mask} with a ResNet-50~\cite{He15} backbone to perform instance segmentation of plant objects, in a binary manner (not differentiating species).
We train for 500 epochs using the settings from \cite{Halstead2021}.
For fully supervised reference models, we use the validation set to select the best performing model.
We use IoU filtering with a threshold of 0.4 as seen in \cite{Halstead2021}.
Further, we use NMS filtering with thresholds determined via validation for fully supervised models.
As validation sets are missing in our semi-supervised datasets, we use the optimum value for \textit{SB20} in \textit{CN20} and vice versa.
To evaluate instance segmentation, we report mean fgIoU over the objects in our datasets similarly to~\cite{Halstead2021}.

\subsubsection{Panoptic system to recover from missing clicks}

For our experiments on missing input clicks, we use the same parameters and network structure as our other panoptic experiments (see \Cref{sec:experiments:panoptic}), but with the center head enabled as described in \Cref{sec:proposed:panoptic_miss}.
For evaluation we use the standard metrics for panoptic segmentation~\cite{Kirillov2019} of panoptic quality (PQ), segmentation quality (SQ), and recognition quality (RQ).
For our use case, which is to recover missing clicks, we are predominantly interested in RQ as it describes the system's ability to correctly identify object instances.

\section{Results}

\begin{table}[!b]
\vspace{-10pt}
\caption{Object segmentation results (mean object IoU) for the different one-click models trained on only 10\% of the instances in our \textit{SB20} and \textit{CN20} datasets.} \label{tab:results:weakseg}
\centering
\begin{tabular}{lrr}
\toprule
Model variant & \textit{SB20} & \textit{CN20} \\
\midrule
Baseline (standard one-click) & 67.7 & 68.0 \\
Additional negative clicks & \textbf{69.5} & \textbf{71.4} \\
Panoptic system & 68.1 & 68.8 \\
\bottomrule
\end{tabular}
\end{table}

\begin{figure}[t!]
    \vspace{8pt}
    \centering
    \includegraphics[width=.48\textwidth]{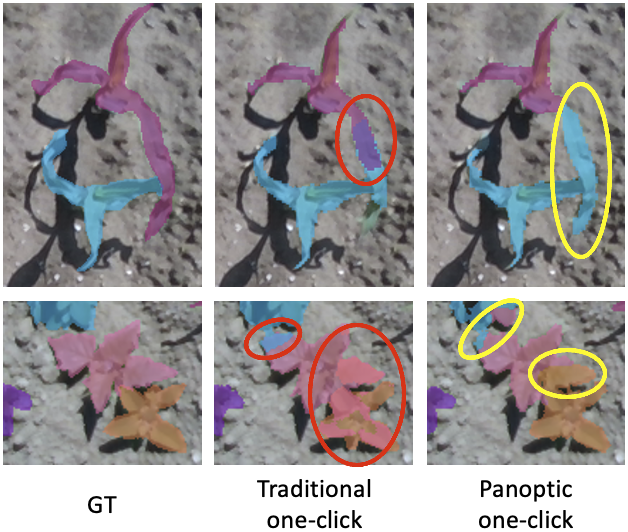}
    \caption{Selected one-click segmentation predictions from our best traditional model (with additional negative clicks; center) or panoptic model (right). Edge cases where both approaches show errors and considerable differences. Red highlights show misclassification errors resulting in overlaps for the traditional system. Yellow highlights show the non-overlapping errors of our panoptic system in the same regions. Image crops have different zoom factors.}
    \label{fig:results:weakseg}
    \vspace{-12pt}
\end{figure}

We perform three sets of experiments.
First, we quantitatively compare traditional one-click segmentation approaches against our proposed one-click panoptic segmentation approach.
Furthermore, we compare the relative training time, highlighting that our panoptic approach can be trained an order of magnitude faster.

The second set of experiments evaluate the one-click segmentation methods, traditional and panoptic, for the task of semi-supervised learning.
Specifically, the one-click approaches are trained using 10\% of the training data and are then used to generate pseudo-labels for the remaining 90\% of the data.
All of this data, 10\% manual annotations and 90\% pseudo-labels, is then used to train an instance-based segmentation network (Mask R-CNN).

The third set of experiments are an ablation study to evaluate the potential of our panoptic one-click segmentation systems to recover from missing clicks.
We compare recognition quality performance of our panoptic system and different model variants trained with varying amounts of input clicks missing.

\subsection{One-click segmentation}
Here, we present results from our first set of experiments regarding general one-click segmentation performance of our models.

\begin{table}[bp]
\vspace{-10pt}
\caption{Instance segmentation results (mean foreground IoU) of Mask R-CNN when trained using 10\% of the data and 100\% of the data compared to the semi-supervised approaches using standard one-click, one-click with negative clicks, and our proposed one-click panoptic approach.}
\label{table:results:mrcnn}
\centering
\begin{tabular}{lrr}
\toprule
Model variant & \textit{SB20} & \textit{CN20} \\
\midrule
10\% data subset & 28.6 & 31.7 \\
Fully supervised & \textbf{42.7} & \textbf{41.9} \\
\midrule
Semi-supervised (standard one-click) & 37.7 & 38.8 \\
Semi-supervised (with negative clicks) & \textbf{40.9} & 38.5 \\
Semi-supervised (panoptic system) & 38.0 & \textbf{39.6} \\
\bottomrule
\end{tabular}
\end{table}

\subsubsection{Quantitative performance}
In \Cref{tab:results:weakseg}, we report one-click segmentation performance (mIoU) for (1) the \textit{standard one-click} baseline, (2) its extension \textit{with negative clicks} and (3) our novel \textit{panoptic one-click method}.
Including the annotations from remaining objects as negative clicks increases baseline performance by 1.8\% and 3.4\% for \textit{SB20} and \textit{CN20}, respectively.
These results highlight that they should be included whenever available if considering a traditional one-click segmentation system.
Our new method shows similar performance to the stardard one-click approach, increasing 0.4\% on \textit{SB20} and 0.8\% on \textit{CN20}.
The competitive performance of the panoptic one-click system when compared to traditional approaches outlines its suitability for generating psuedo-labels for semi-supervised learning.

\subsubsection{Qualitative comparison}
We demonstrate some qualitative differences between the presented approaches with prediction examples from both datasets in \Cref{fig:results:weakseg}.
We outline figures from the panoptic one-click and traditional negative clicks systems, with particular attention being paid to edge cases which show different types of prediction errors for both systems.
One benefit of the panoptic approach over the traditional approach is that panoptic vision inherently ensures no overlap between objects.
This is not the case for standard methods, which we show for some regions with overlapping misclassification errors highlighted in red.
Non-overlapping errors in the according regions for our panoptic system are further highlighted in yellow.

\subsubsection{Computational performance}
In our setup with a single GPU, we achieved training times of 55 minutes and 38 minutes with our panoptic one-click system on \textit{SB20} and \textit{CN20}, respectively, compared to 674 minutes and 448 minutes with the traditional one-click model with negative clicks.
For both datasets, \textit{SB20} and \textit{CN20}, this is a speed improvement of approximately a factor of 12 which means that our proposed panoptic one-click system is an order of magnitude faster to train.

\subsection{Semi-supervised instance segmentation}

In this experiment, we evaluate the performance of the generated pseudo-labels in a Mask R-CNN based semi-supervised system.
We evaluate this performance on five different systems: (1) using our 10\% data split; (2) fully supervised; (3) semi-supervised standard one-click; (4) semi-supervised one-click with negative clicks; and (5) our novel semi-supervised panoptic one-click.

\Cref{table:results:mrcnn} outlines the instance segmentation performance of these five systems.
We report the instance-wise segmentation performance using mean foreground IoU.
As expected, using just 10\% of the data results in the poorest performance across both datasets, and the fully supervised system performs best.
We use the fully supervised approach as our upper reference score and expect a good performing semi-supervised approach to obtain values commensurate with it.

When considering our semi-supervised approaches we see a consistent improvement over the 10\% data subsets.
On \textit{SB20} the negative clicks system yields close to fully supervised performance (40.9 versus 42.7).
Standard one-click performs slightly lower at 37.7, similar to our panoptic system at 38.0 which is a considerable 9.4 point increase when compared to not using click labels.
On \textit{CN20}, standard one-click and negative clicks models perform on the same level (38.8 and 38.5).
Our panoptic model at 39.6 gets closest to the fully supervised reference model with only a 2.3 point difference while outperforming the 10\% data subset by 7.9 points.

Finally, while the proposed panoptic approach does not consistently perform better than the traditional approach the speed benefits (better than an order of magnitude faster) mean that when rapid prototyping is required this approach is beneficial.
An additional potential benefit of using a panoptic system is that it can recover when clicks are missing, we evaluate this in the following ablation study.

\subsection{Ablation study - Recovering from missing input clicks}

\begin{table}[b!]
\caption{Panoptic segmentation performance with different percentages of missing input clicks of \textit{SB20} data.
First line is our original click segmentation model with input clicks used for post processing panoptic segmentation results.
Remaining models use the original scheme from \cite{Cheng2020} to detect centers using the center head.}
\label{tab:results:missing}
\centering
\begin{tabular}{lrrr}
\toprule
Percentage of Clicks Missing &    PQ &    SQ &    RQ \\
\midrule
0\% [centers provided by user (clicks)] &  75.0 &  85.0 &  86.0 \\
\midrule
0\% [centers predicted by network]  &  74.0 &  84.9 &  84.8 \\
25\%                      &  69.8 &  84.0 &  79.8 \\
50\%                     &  70.1 &  84.4 &  79.8 \\
75\%                     &  69.2 &  83.9 &  78.9 \\
100\%                     &  66.4 &  84.5 &  74.5 \\
\bottomrule
\end{tabular}
\end{table}

In these experiments we evaluate the potential for a panoptic one-click system to recover when clicks are missing.
For these experiments we use the available clicks as an extra input channel but we do not use them as the click locations for the output of the network.
Instead, we train the panoptic network to estimate the click locations and use these to derive the per-object pixel-wise segmentation map.

The results in \Cref{tab:results:missing} highlight that the panoptic one-click system has the potential to recover from missing clicks.
It can be seen, in the top two lines in the table, that using the panoptic one-click network to  predict the object center locations results in only slightly worse performance (in terms of RQ) than using the user input clicks.
The RQ performance drops by only 1.2 points going from 86.0 to 84.8 for user provided center locations and network predicted locations respectively.
When user clicks are removed (or missed) the RQ performance drops to 79.8 and remains at this level until 75\% of the clicks are removed (or missed).
With 75\% of the clicks missing the RQ performance degrades by only 0.9 points and when no clicks are given (100\% missing clicks) the system still has an RQ performance of 74.5.
This indicates that the panoptic system is able to well estimate the object locations, even when large parts (50\% or 75\%) of the input clicks are missing.

\section{Conclusion}

In this paper we have presented panoptic one-click segmentation and showed that it is a suitable tool to produce pseudo-labels from minimal user input (single click per object).
Our approach is approximately an order of magnitude (12 times) faster than the traditional one-click approaches while achieving IoU values 68.1 and 68.8 on two challenging arable farmland datasets, commensurate with traditional approaches.
We have demonstrated that our method can be used to drastically reduce the labelling effort when creating novel datasets while maintaining accuracy.
To outline the practicality of our panoptic one-click segmentation approach we show its effectiveness at generating two instance-based semantic segmentation datasets.
We use only 10\% of the fully annotated data and augment the final 90\% with our generated pseudo-labels (generated from single clicks), creating a novel dataset for evaluating Mask R-CNN.
On this semi-supervised experiment we are able to improve IoU scores from 28.6 to 38 for \textit{SB20} and 31.7 to 39.6 for \textit{CN20}, exploiting our single click pseudo-labels.
Finally, we explore the potential of panoptic one-click segmentation to recover clicks that are missing during manual annotation.
We show that even when 75\% of the clicks are missing our approach is still able to achieve a recognition quality score of 78.9. 
This outlines our approach's ability to recover from missing clicks, something that traditional one-click approaches can not achieve.

\bibliography{refs}
\bibliographystyle{plain}

\end{document}